%% file: paper.tex
\documentclass[acmsmall]{acmart}
\renewcommand\footnotetextcopyrightpermission[1]{} 
\AtBeginDocument{%
  \providecommand\BibTeX{{%
    \normalfont B\kern-0.5em{\scshape i\kern-0.25em b}\kern-0.8em\TeX}}}

\acmJournal{TIST}

\usepackage{graphicx}
\usepackage{todonotes}
\usepackage[caption=false,font=footnotesize]{subfig}
\usepackage{tabularx, booktabs}
\usepackage{algpseudocode}
\usepackage{algorithmicx}

\usepackage{color}
\definecolor{Orange}{rgb}{0.9,0.5,0}
\definecolor{NavyBlue}{rgb}{0.1, 0.4, 0.8}
\definecolor{Magenta}{rgb}{0.8, 0.1, 0.6}
\definecolor{Green}{rgb}{0.1, 0.8, 0.3}
\definecolor{DarkGreen}{rgb}{0.0, 0.7, 0.2}
\definecolor{Brown}{rgb}{0.4, 0.3, 0.1}
\definecolor{Burgundy}{rgb}{0.5, 0.0, 0.13}
\definecolor{BrightCerulean}{rgb}{0.11, 0.67, 0.84}
\definecolor{BlueViolet}{rgb}{0.33,0.1,0.5}

\begin{document}

\def\etal{\emph{et al.}}
\def\ie{\emph{i.e.}}
\def\eg{\emph{e.g.}}
\title{Semi-supervised Federated Learning for Activity Recognition}

\author{Yuchen Zhao} 
\email{yuchen.zhao19@imperial.ac.uk}
\author{Hanyang Liu}
\email{hanyang.liu18@imperial.ac.uk}
\author{Honglin Li}
\email{honglin.li20@imperial.ac.uk}
\author{Payam Barnaghi}
\email{p.barnaghi@imperial.ac.uk}
\author{\\Hamed Haddadi}
\email{h.haddadi@imperial.ac.uk}
\affiliation{
  \institution{Imperial College London}
  \country{UK}
}

\renewcommand{\shortauthors}{Yuchen Zhao, Hanyang Liu, Honglin Li, Payam Barnaghi, Hamed Haddadi}

\begin{abstract}
Training deep learning models on in-home IoT sensory data is commonly used to
recognise human activities. Recently, federated learning systems that use
edge devices as clients to support local human activity recognition have
emerged as a new paradigm to combine local (individual-level) and global
(group-level) models. This approach provides better scalability and
generalisability and also offers better privacy compared with the traditional
centralised analysis and learning models. The assumption behind federated
learning, however, relies on supervised learning on clients. This requires a
large volume of labelled data, which is difficult to collect in uncontrolled
IoT environments such as remote in-home monitoring.

In this paper, we propose an activity recognition system that uses
semi-supervised federated learning, wherein clients conduct unsupervised
learning on autoencoders with unlabelled local data to learn general
representations, and a cloud server conducts supervised learning on an
activity classifier with labelled data. Our experimental results show that
using a long short-term memory autoencoder and a Softmax classifier, the
accuracy of our proposed system is higher than that of both centralised
systems and semi-supervised federated learning using data augmentation. The
accuracy is also comparable to that of supervised federated learning systems.
Meanwhile, we demonstrate that our system can reduce the number of needed
labels and the size of local models, and has faster local activity
recognition speed than supervised federated learning does.
\end{abstract}



\pagestyle{plain} 
\maketitle

\input{intro}
\input{related}
\input{method}
\input{evaluation}
\input{results}
\input{discussion}
\input{conclusions}

\bibliographystyle{ACM-Reference-Format}
\bibliography{paper}

\end{document}

%% file: intro.tex
\section{Introduction}
Modern smart homes are integrating more and more Internet of Things (IoT)
technologies in different application scenarios. The IoT devices can collect
a variety of time-series data, including ambient data such as occupancy,
temperature, and brightness, and physiological data such as weight and blood
pressure. With the help of machine learning (ML) algorithms, these sensory
data can be used to recognise people's activities at home. Human activity
recognition (HAR) using IoT data has the promise that can significantly
improve quality of life for people who require in-home care and support. For
example, anomaly detection based on recognised activities can raise alerts
when an individual's health deteriorates. The alerts can then be used for
early interventions ~\cite{Enshaeifar2018:machine-learning,
Cao2015,Queralta2019}. Analysis of long-term activities can help identify
behaviour changes, which can be used to support clinical decisions and
healthcare plans~\cite{Enshaeifar2018:iot}.

An architecture for HAR is to deploy devices with computational resources at
the edge of networks, which is normally within people's homes. Such ``edge
devices'' are capable of communicating with sensory devices to collect and
aggregate the sensory data, and running ML algorithms to process the in-home
activity and movement data. With the help of a cloud back-end, these edge
devices can form a federated learning (FL) system~\cite{McMahan2016,Yang2019,
Zhao2020:edgesys}, which is increasingly used as a new system to learn at
population-level while constructing personalised edge models for HAR. In an
FL system, clients jointly train a global Deep Neural Network (DNN) model by
sharing their local models with a cloud back-end. This design enables clients
to use their data to contribute to the training of the model without
breaching privacy. One of the assumptions behind using the canonical FL
system for HAR is that data on clients are labelled with corresponding
activities so that the clients can use these data to train supervised local
DNN models. In HAR using IoT data, due to the large amount of time-series
data that are continuously generated from different sensors, it is difficult
to guarantee that end-users are capable of labelling activity data at a large
scale. Thus, the availability of labelled data on clients is one of the
challenges that impede the adoption of FL systems in real-world HAR
applications.

Existing solutions to utilise unlabelled data in FL systems is through data
augmentation~\cite{jeong2020federated,liu2020rc, zhang2020benchmarking}. The
server of an FL system keeps some labelled data and use them to train
a global model through supervised learning. The clients of the system receive the
global model and use it to generate pseudo labels on augmented local data.
However, this approach couples the local training on clients with the
specific task (\ie, labels) from the server. If a client accesses multiple
FL servers for different tasks, it has to generate pseudo labels for each of
them locally, which increases the cost of local training.

In centralised ML, unsupervised learning on DNN such as
autoencoders~\cite{Baldi2011} has been widely used to learn general
representations from unlabelled data. The learned representations can then be
utilised to facilitate supervised learning models with labelled data. A
recent study by van Berlo \etal~\cite{vanBerlo2020:edgesys} shows that
temporal convolutional networks can be used as autoencoders to learn
representations on clients of an FL system. The representations can help with
training of the global supervised model of an FL system. The resulting
model's performance is comparable to that of a fully supervised algorithm.
Building upon this promising result, we propose a semi-supervised FL system
that realises activity recognition using time-series data at the edge,
without labelled IoT sensory data on clients, and evaluate how different
factors (e.g., choices of models, the number of labels, and the size of
representations) affect the performance (e.g., accuracy and inference time)
of the system.

In our proposed design, clients locally train autoencoders with unlabelled
time-series sensory data to learn representations. These local autoencoders
are then sent to a cloud server that aggregates them into a global
autoencoder. The server integrates the resulting global autoencoder into the
pipeline of the supervised learning process. It uses the encoder component of
the global autoencoder to transform a labelled dataset into labelled
representations, with which a classifier can be trained. Such a labelled
dataset on the cloud back-end server can be provided by service providers
without necessarily using any personal data from users (e.g., open data or
data collected from laboratory trials with consents). Whenever the server
selects a number of clients, both the global autoencoder and the global
classifier are sent to the clients to support local activity recognition.

We evaluated our system through simulations on different HAR datasets, with
different system component designs and data generation strategies. We also
tested the local activity recognition part of our system on a Raspberry Pi 4
model B, which is a low-cost edge device. With the focus on HAR using
time-series sensory data, we are interested in answering the research
questions as follows:

\begin{itemize}
    \item Q1. How does semi-supervised FL using autoencoders perform in
    comparison to supervised learning on a centralised server?
    \item Q2. How does semi-supervised FL using autoencoders perform in
    comparison to semi-supervised FL using data augmentation?
    \item Q3. How does semi-supervised FL using autoencoders perform in
    comparison to supervised FL?
    \item Q4. How do the key parameters of semi-supervised FL, including the
    number of labels on the server and the size of learned representations,
    affect its performance.
    \item Q5. How efficient is semi-supervised FL on low-cost edge
    devices?
\end{itemize}

Our experimental results demonstrate several key findings:
\begin{itemize}
    \item Using long short-term memory autoencoders as local models and a
    Softmax classifier model as a global classifier, the accuracy of our
    system is higher than that of a centralised system that only conducts
    supervised learning in the cloud, which means that learning general
    representations locally improves the performance of the system.
    \item Our system also has higher accuracy than semi-supervised FL using
    data augmentation to generate pseudo labels does.
    \item Our system can achieve comparable accuracy to that of a supervised
    FL system.
    \item By only conducting supervised learning in the cloud, our system can
    significantly reduce the needed number of labels without losing much accuracy.
    \item By using autoencoders, our system can reduce the size
    of local models. This can potentially contribute to the reduction of
    upload traffic from the clients to the server.
    \item The processing time of our system when recognising activities on a
    low-cost edge device is acceptable for real-time applications and is
    significantly lower than that of supervised FL.
\end{itemize}

%% file: related.tex
\section{Related}
As one of the key applications of IoT that can significantly improve the quality
of people's lives, HAR has attracted an enormous amount of research. Many HAR
systems have been proposed to be deployed at the edge of networks, thanks to
the evergrowing computational power of different types of edge devices.

\subsection{HAR at the edge}
In comparison to having both data and algorithms in the cloud, edge
computing~\cite{Shi2016} instead deploys devices closer to end users of
services, which means that data generated by the users and computation on
these data can stay on the devices locally. Modern edge devices such as Intel
Next Unit of Computing (NUC)~\cite{nuc} and Raspberry Pi~\cite{pi} are
capable of running DNN models~\cite{Servia-Rodriguez2018, Chen2019} and
providing real-time activity recognition~\cite{Liu2018, Cartas2019} from
videos. Many deep learning models such as long short-term memory
(LSTM)~\cite{lstm,Guan2017, Hammerla2016} or convolutional neural network
(CNN) ~\cite{Hammerla2016} can be applied at the edge for HAR. For example,
Zhang \etal~\cite{Zhang2018} proposed an HAR system that utilised both edge
computing and back-end cloud computing. One implementation of this kind of HAR
edge systems was proposed by Cao \etal~\cite{Cao2015}, which implemented fall
detection both at the edge and in the cloud. Their results show that their
system has lower response latency than that of a cloud based system. Queralta
\etal~\cite{Queralta2019} also proposed a fall detection system that achieved
over 90\% precision and recall. Uddin~\cite{Uddin2019} proposed a system that
used more diverse body sensory data including electrocardiography (ECG),
magnetometer, accelerometer, and gyroscope readings for activity recognition.

These HAR systems, however, send the personal data of their users to a
back-end cloud server to train deep learning models, which poses great privacy
threats to the data subjects. Servia-Rodr\'{i}guez
\etal~\cite{Servia-Rodriguez2018} proposed a system in which a small group of
users voluntarily share their data to the cloud to train a model. Other users
in the system can download this model for local training, which protects the
privacy of the majority in the system but does not utilise the fine trained
local models from different users to improve the performance of each other's
models. To improve the utility of local models and protect privacy at the
same time, we apply federated learning~\cite{McMahan2016} to HAR at the edge,
which can train a global deep learning model with constant contributions from
users but does not require the users to send their personal data to the
cloud.

\subsection{HAR with federated learning}
Federated learning (FL)~\cite{McMahan2016,Yang2019} was proposed as an
alternative to traditional cloud based deep learning systems. It uses a cloud
server to coordinate different clients to collaboratively train a global
model. The server periodically sends the global model to a selection of
clients that use their local data to update the global model.
The resulting local models from the clients will be sent back to the server
and be aggregated into a new global model. By this means, the global model is
constantly updated using users' personal data, without having these data in
the server. Since FL was proposed, it has been widely adopted in many
applications~\cite{Li2020:CIE, Yu2020} including HAR. Sozinov
\etal~\cite{Sozinov2018} proposed an FL based HAR system and they demonstrated
that its performance is comparable to that of its centralised counterpart,
which suffers from privacy issues. Zhao \etal~\cite{Zhao2020:edgesys}
proposed an FL based HAR system for activity and health monitoring. Their
experimental results show that, apart from acceptable accuracy, the inference
time of such a system on low-cost edge devices such as Raspberry Pi is
marginal. Feng \etal~\cite{Feng2020} introduced locally personalised models
in FL based HAR systems to further improve the accuracy for mobility
prediction. Specifically, HAR applications that need both utility and privacy
guarantees such as smart healthcare can benefit from the accurate recognition
and the default privacy by design of FL. For example, the system recently
proposed by Chen \etal~\cite{Chen2020} applied FL to wearable healthcare,
with a specific focus on the auxiliary diagnosis of Parkinson's disease.

Existing HAR systems with canonical FL use supervised learning that relies on
the assumption that all local data on clients are properly labelled with
activities. This assumption is difficult to be satisfied in the scenario of
IoT using sensory data. Compared to the existing FL based HAR systems, we aim
to address this issue by utilising semi-supervised machine learning, which
does not need locally labelled data.

\subsection{Semi-supervised federated learning}
Semi-supervised learning combines both supervised learning that requires
labelled data and unsupervised learning that does not use labels when
training DNN models. Traditional centralised ML has benefited from
semi-supervised learning techniques such as transfer learning
~\cite{Khan2018, Zhang2019} and autoencoders~\cite{Baldi2011}. These
techniques have been widely used in centralised ML such as learning
time-series representations from videos~\cite{Srivastava2015}, learning
representations to compress local data~\cite{Hu2020}, and learning
representations that do not contain sensitive
information~\cite{Malekzadeh2018}.

The challenge of having available local labels in FL has motivated a number of
systems that aim to realise FL in a semi-supervised or self-supervised fashion.
The majority of the existing solutions in this area focuses on generating pseudo
labels for unlabelled data and using these labels to conduct supervised
learning~\cite{jeong2020federated,liu2020rc, zhang2020benchmarking,
long2020fedsemi, zhang2021106679, kang2020fedmvt, wang2020graphfl,
yang2020federated}. For example, Jeong~\etal~\cite{jeong2020federated} use data
augmentation to generate fake labels and keep the consistency of the labels
across different FL clients. However, the inter-client consistency requires
some clients to share their data with others, which poses privacy issues.
Liu~\etal~\cite{liu2020rc} use labelled data on an FL server to train a model
through supervised learning and then send this model to FL clients to
generate labels on their local data. These solutions couple the local
training on clients with the specific task from the server, which means that
a client has to generate pseudo labels for all the servers that have different
tasks.

Another direction of semi-supervised FL is to conduct unsupervised learning
on autoencoders locally on clients instead of generating pseudo labels.
Compared with existing solutions, the trained autoencoders learn general
representations from data, which are independent from specific tasks.
Preliminary results from the work by van Berlo
~\etal~\cite{vanBerlo2020:edgesys} show promising potential of using
autoencoders to implement semi-supervised FL. Compared to their work, we
evaluate different local models (\ie, autoencoders, convolutional
autoencoders, and LSTM autoencoders), investigate different design
considerations, and test how efficient its local activity recognition is when
running on low-cost edge devices.

%% file: method.tex
\section{Methodology}
Our goal is to implement HAR using an FL system, without having any labelled
data on the edge clients. We first introduce the long short-term memory
model~\cite{lstm}, which is a technique for analysing time-series data for
HAR. We then introduce autoencoders, which are the key technique for deep
unsupervised learning. We finally demonstrate the design of our proposed
semi-supervised FL system and describe how unsupervised and supervised
learning models are used in our framework.

\subsection{Long short-term memory}
\label{sec:lstm}
\begin{figure}[t!]
    \includegraphics[width=.75\linewidth]{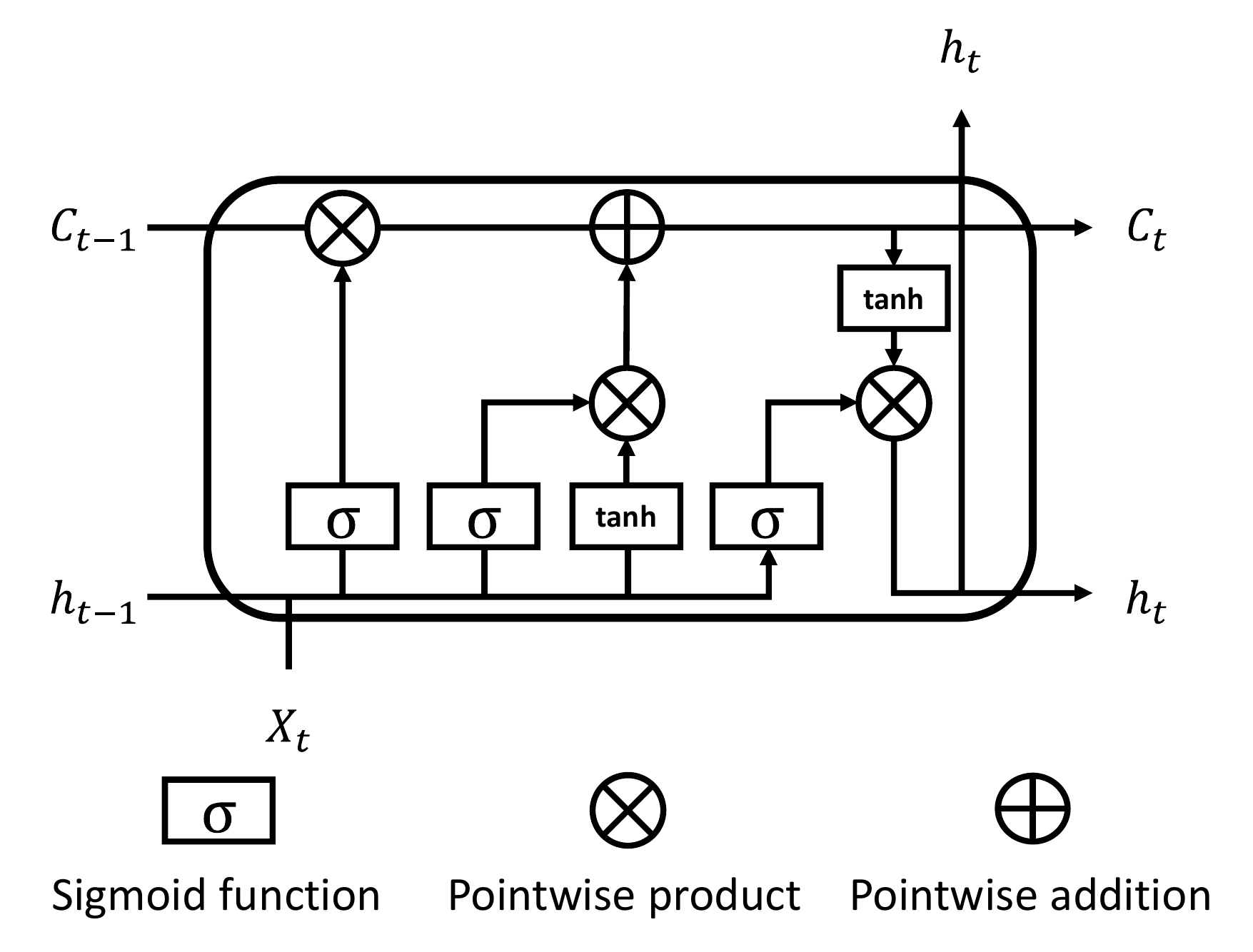}
    \caption{Network structure of a long short-term memory (LSTM) cell. At each
    time point $t$, the current cell state $C_{t}$ and hidden state $h_{t}$ is
    dependent on the previous cell state $C_{t-1}$, the previous hidden state
    $h_{t}$, and the current observed data point $X_{t}$.}
    \label{fig:lstm}
\end{figure}

The long short-term memory (LSTM) belongs to recurrent neural network (RNN)
models, which are a class of DNN that processes sequences of data points such
as time-series data. At each time point of the time series, the output of an
RNN, which is referred to as the ``hidden state'', is fed to the network
together with the next data point in the time-series sequence. An RNN works
in a way that, as time proceeds, it recurrently takes and processes the
current input and the previous output (\ie, the hidden state), and generates
a new output for the current time. Specifically for LSTM,
Fig.~\ref{fig:lstm} shows the network structure of a basic LSTM unit, which
is called an LSTM \emph{cell}. At each time $t$, it takes three input
variables, which are the current observed data point $X_{t}$, the previous
state of the cell $C_{t-1}$, and the previous hidden state $h_{t-1}$. For the
case of applying LSTM to HAR, $X_{t}$ is a vector of all the observed
sensory readings at time $t$. $h_{t}$ is the hidden state of the activity to
be recognised in question.

LSTM can be used in both supervised learning and unsupervised learning. For
supervised learning, each $X_{t}$ of a time-series sequence has a
corresponding label $Y_{t}$ (\eg, activity class at time point $t$) as the ground
truth. The hidden state $h_{t}$ can be fed into a ``Softmax classifier'' that
contains a fully-connected layer and a Softmax layer. By this means, such an
LSTM classifier can be trained against the labelled activities through
feedforward and backpropagation to minimise the loss (\eg, Cross-entropy
loss) between the classifications and the ground truth. For unsupervised
learning, LSTM can be trained as components of an autoencoder, which we will
describe in detail in Sec.~\ref{sec:autoencoder}.

\subsection{Autoencoder}
\label{sec:autoencoder}
\begin{figure}[t!]
    \subfloat[][Autoencoder with fully connected layers]{\includegraphics[width=.45\linewidth]{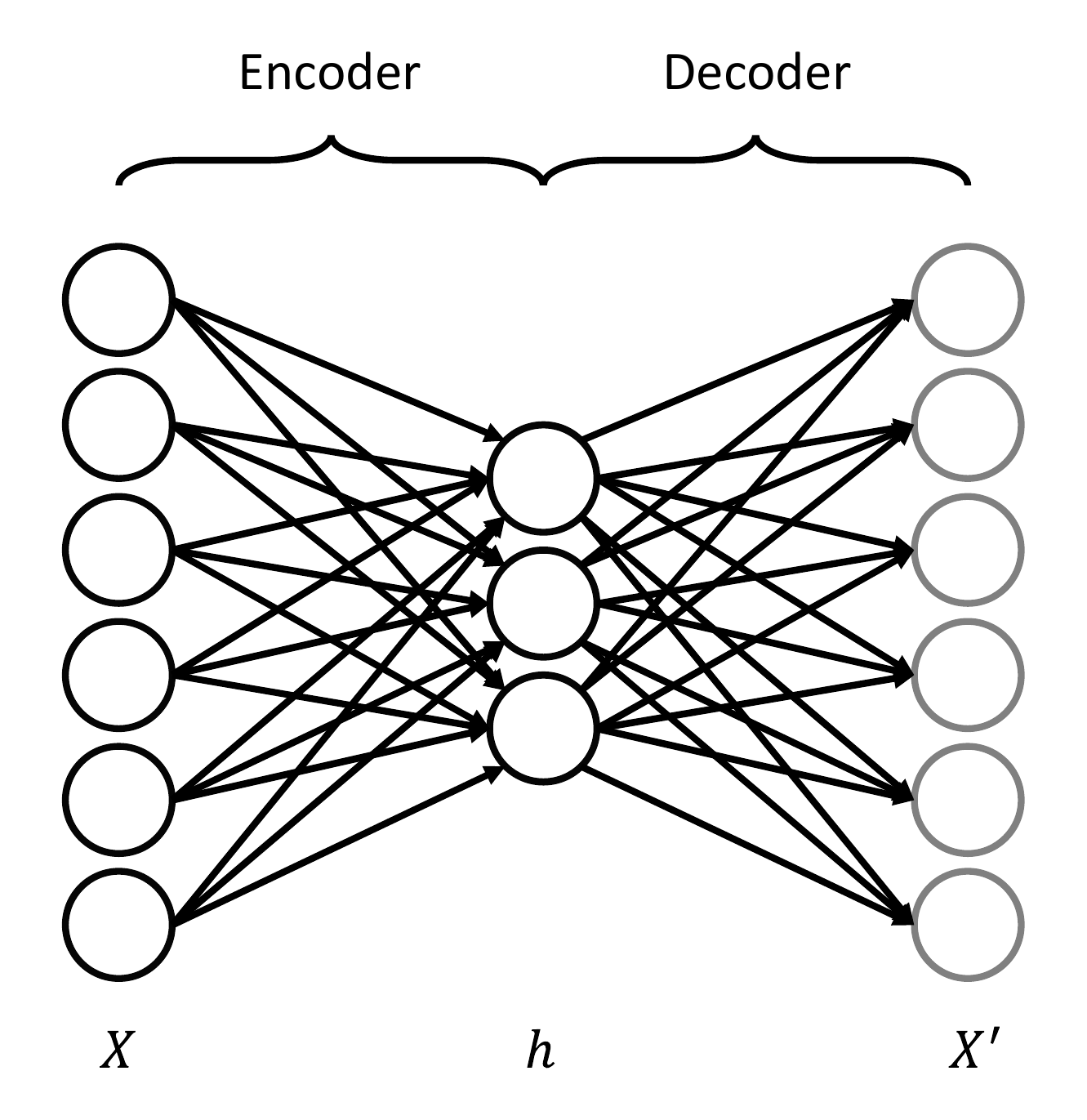}\label{sub_fig:ae}}
    \subfloat[][Autoencoder with concolutional layers]{\includegraphics[width=.45\linewidth]{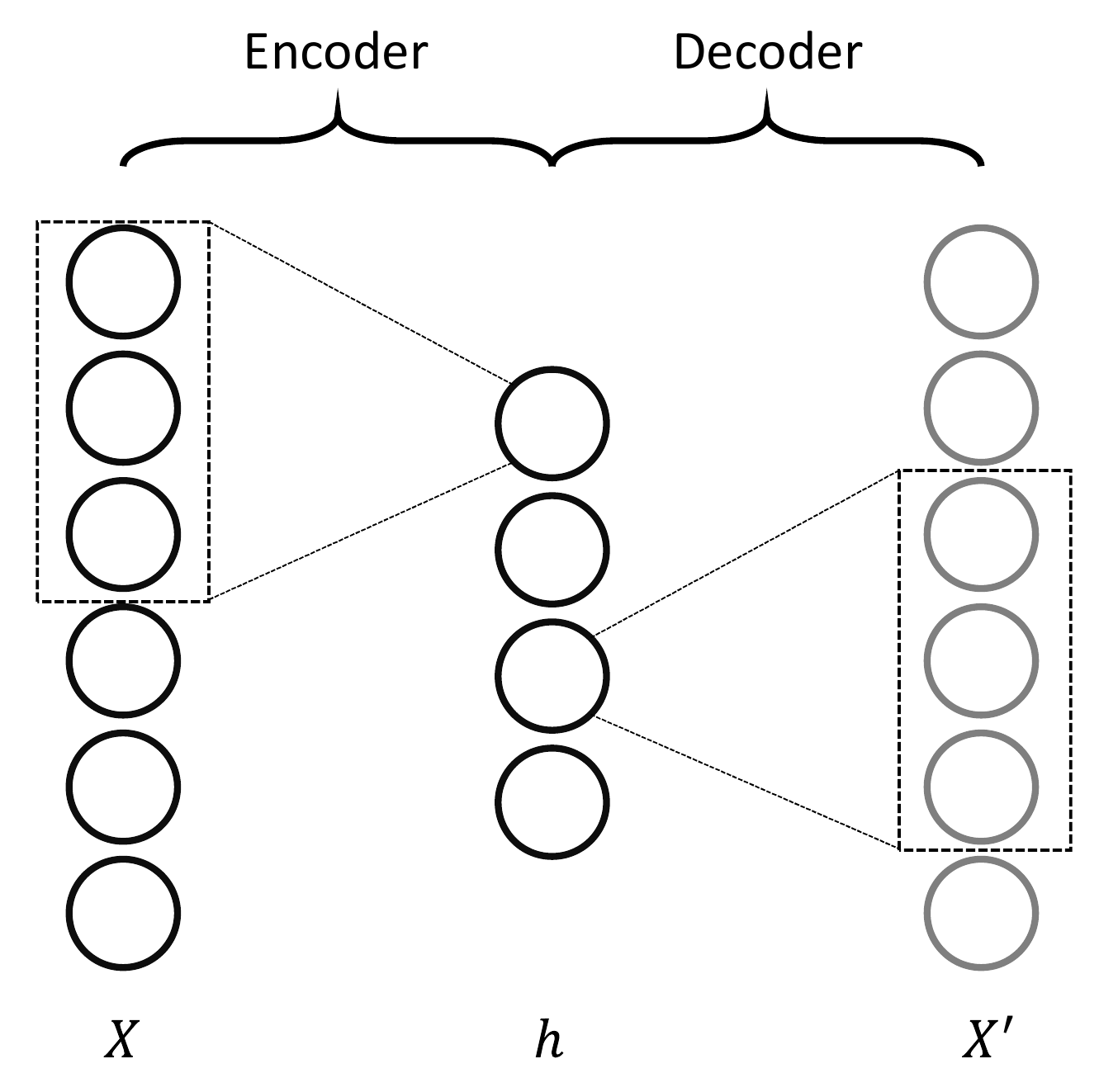}\label{sub_fig:cae}}
    \caption{Network structures of a simple autoencoder and a convolutional
    autoencoder. The encoder part compresses the input $X$ into a
    representation $h$ that has fewer dimensions. The decoder part tries to
    generate a reconstruction $X^\prime$ from $h$, which is supposed to be
    close to $X$.}
    \label{fig:ae}
\end{figure}

An autoencoder~\cite{Baldi2011} is a type of neural network that is used to
learn latent feature representations from data. Different from supervised
learning that aims to learn a function $f(X)\rightarrow Y$ from input
variables $X$ to labels $Y$, an autoencoder used in unsupervised learning
tries to \emph{encode} $X$ to its latent representation $h$ and to
\emph{decode} $h$ into a reconstruction of $X$, which is presented as
$X^\prime$. Fig.~\ref{fig:ae} demonstrates two types of autoencoders that use
different neural networks. The simple autoencoder in
Fig.~\ref{sub@sub_fig:ae} uses fully connected layers to encode $X$ into
$h$ and then to decode $h$ into $X^\prime$. The convolutional autoencoder in
Fig.~\ref{sub@sub_fig:cae} uses a convolutional layer that moves small
kernels alongside the input $X$ and conducts convolution operations on each
part of $X$ to encode it into $h$. The decoder part uses a transposed
convolutional layer that moves small kernels on $h$ to upsample it into
$X^\prime$.

Ideally, $X^\prime$ is supposed to be as close to $X$ as possible, based on
the assumption that the key representations of $X$ can be learned and encoded
as $h$. As the dimensionality of $h$ is lower than that of $X$, there is less
information in $h$ than in $X$. Thus the reconstructed $X^\prime$ is likely
to be a distorted version of $X$. The goal of training an autoencoder is to
minimise the distortion, \ie, minimising a loss function $L(X,X^\prime)$,
thereby producing an encoder (e.g., fully connected hidden layers or
convolutional hidden layers) that can capture $X$'s most useful information
in its representation $h$.

\begin{figure}[t!]
   \includegraphics[width=.75\linewidth]{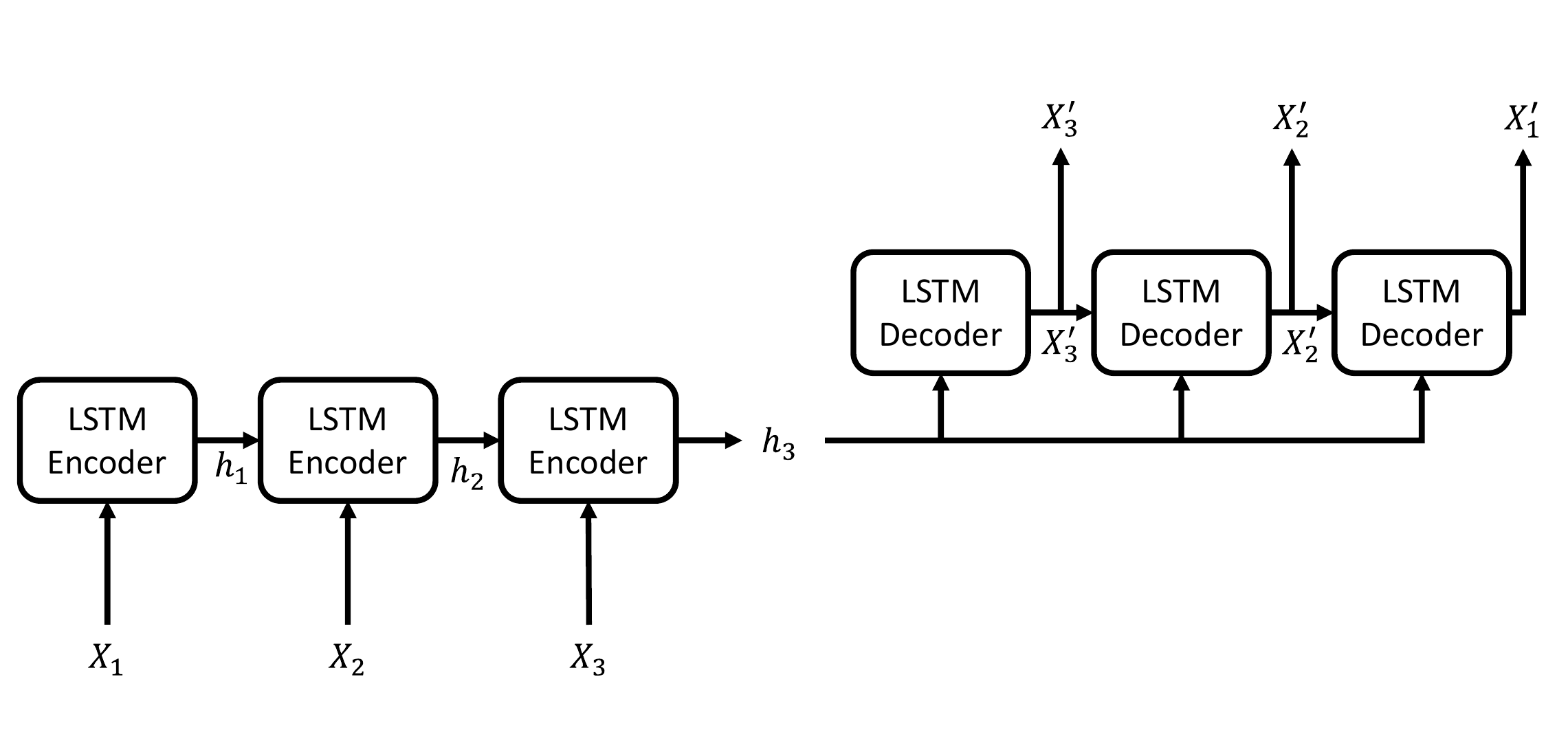} 
   \caption{Network structure of an LSTM-autoencoder. The time-series
   sequence $(X_{1},X_{2},X_{3})$ is input into an LSTM encoder cell and the final output hidden state $h_{3}$ (\ie, after $X_{3}$ is input into the
   encoder) is the representation of $X_{3}$ in the context of $(X_{1}, X_{2},
   X_{3})$. A sequence of the learned representation with the same length as that of the original sequence, \ie, $(h_{3},h_{3},h_{3})$, is input into
   an LSTM decoder cell. The output sequence tries to reconstruct the
   original sequence in reversed order.}
   \label{fig:lstm-ae}
\end{figure}

As mentioned Sec.~\ref{sec:lstm}, LSTM can also be used as components of an
LSTM-autoencoder~\cite{Srivastava2015} to encode time-series data. As shown
in Fig.~\ref{fig:lstm-ae}, an LSTM cell is used as the encoder of an
autoencoder and takes a time-series sequence as its input. The final hidden
state $h_{3}$ is the representation of $X_{3}$ in the context of the sequence
$(X_{1},X_{2},X_{3})$. As the hidden state of the LSTM encoder is based on
both the input observation and the previous hidden states, the representation
generated in this way compresses information of both the features in the
observation and the time-series sequence. The decoder, which is another LSTM
cell, reconstructs the original sequence in a reversed order. Thus, the goal
of an LSTM-autoencoder is to minimise the loss between the original and the
reconstructed sequences.

Since our system runs unsupervised learning locally at the edge and
supervised learning in the cloud, we consider simple autoencoders,
convolutional autoencoders, and LSTM-autoencoders in our proposed system, in
order to understand how the location where time-series information is
captured (\ie, in supervised learning or unsupervised learning) affect the
performance of our system.

\subsection{System design}
\begin{figure}[t!]
    \includegraphics[width=.7\linewidth]{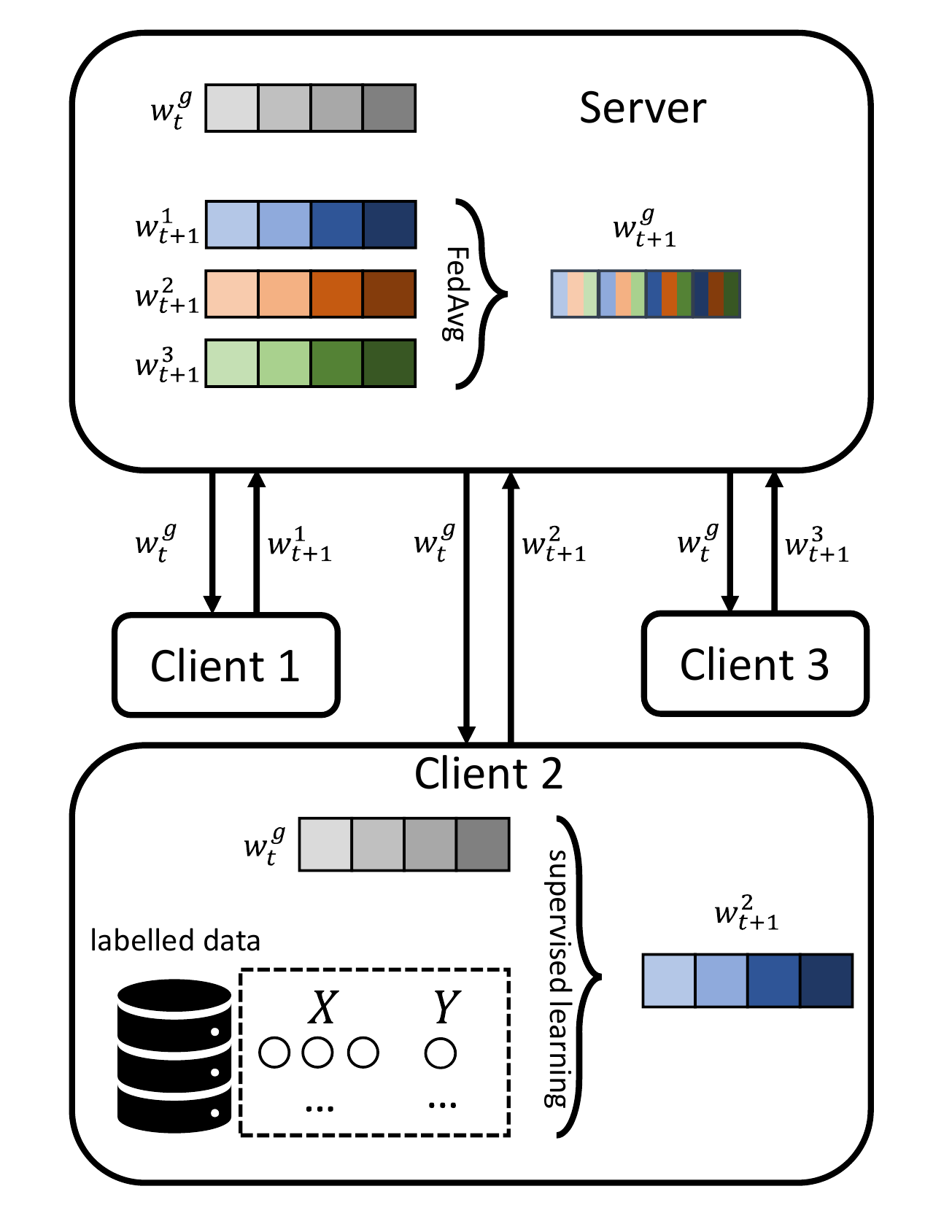}
    \caption{System structure of a canonical federated learning (FL) system with supervised learning. The server selects 3 clients and sends the global $w^{g}_{t}$ to them and the clients use their labelled data to 
   update $w^{g}_{t}$ into their local models, which are then sent to the server to be aggregated into a new global model using the FedAvg algorithm.}
    \label{fig:system-supervised}
\end{figure}

In a canonical FL system, as in a client-server structure, a cloud server
periodically sends a global model to selected clients for updating the model
locally. As shown in Fig.~\ref{fig:system-supervised}, in each communication
round $t$, a global model $w^{g}_{t}$ is sent to three selected clients,
which conduct supervised learning on $w^{g}_{t}$ with their labelled local
data. The resulting local models are then sent to the server, which uses the
federated averaging (FedAvg) algorithm~\cite{McMahan2016} to aggregate these
models into a new global model $w^{g}_{t+1}$. The server and clients repeat
this procedure through multiple communication rounds between them, thereby
fitting the global model to clients' local data without releasing the data to
the server.

In order to address the lack of labels on clients in HAR with IoT sensory
data, our proposed system applies semi-supervised learning in an FL system,
in which clients use unsupervised learning to train autoencoders with their
unlabelled data, and a server uses supervised learning to train a classifier
that can map encoded representations to activities with a labelled dataset.

\begin{figure*}[t!]
    \includegraphics[width=\linewidth]{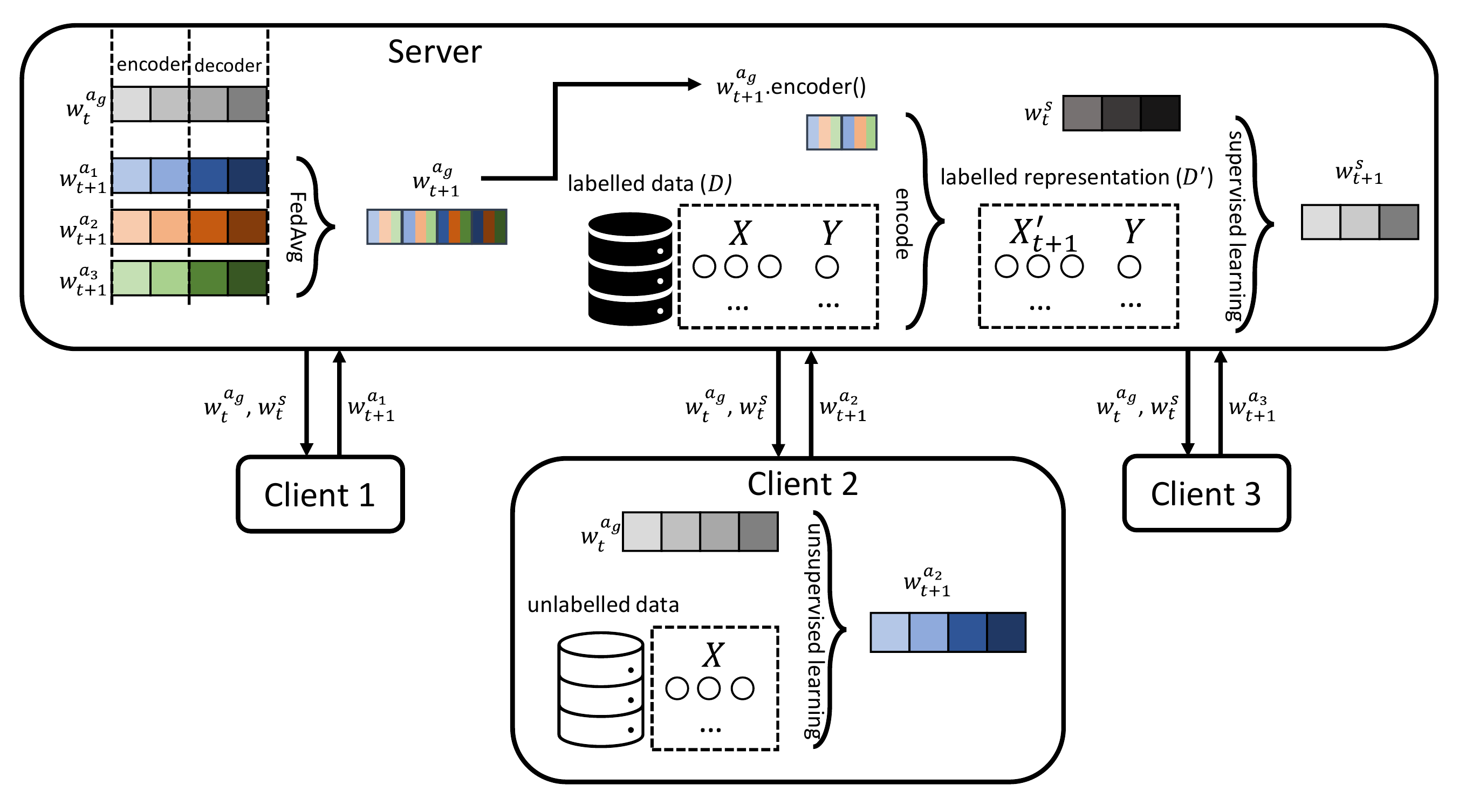}
    \caption{System structure of our semi-supervised FL system. A server and three clients follow the standard FL procedure to update a global
    autoencoder $w^{a_{g}}_{t}$ to $w^{a_{g}}_{t+1}$. Then the server uses
    the encoder of $w^{a_{g}}_{t+1}$ to encode the labelled dataset $D$ on the
    server into a labelled representation dataset $D^\prime$. It then conducts
    supervised learning on a classifier $w^{s}_{t}$ with $D^\prime$ and
    generates a new classifier $w^{s}_{t+1}$.}
    \label{fig:system-semi}
\end{figure*}

As shown in Fig.~\ref{fig:system-semi}, in each communication round, the
server sends a global autoencoder $w^{a_{g}}_{t}$ to selected clients. In
order to update $w^{a_{g}}_{t}$ locally, clients run unsupervised
learning on $w^{a_{g}}_{t}$ with their unlabelled local data and then send
the resulting local autoencoders to the server. The server follows the
standard FedAvg algorithm to generate a new global autoencoder
$w^{a_{g}}_{t+1}$, which is then plugged into the pipeline of supervised
learning with a labelled dataset $D=(X,Y)$. The server first uses the encoder
part of $w^{a_{g}}_{t+1}$ to encode the original features $X$ into
representations $X^\prime$ in order to generate a labelled representation
dataset $D^\prime=(X^\prime,Y)$. Then the server conducts supervised learning
with $D^\prime$ to update a classifier $w^{s}_{t}$ into $w^{s}_{t+1}$.
Fig.~\ref{fig:alg} shows the detailed semi-supervised algorithm of our
system.

\begin{figure}
    \hrule
    \vspace{0.1cm}
    \begin{algorithmic}[1]
        \Require $K$: number of clients; $C$: fraction of clients; $D=(X,Y)$:
        labelled dataset in the format as $(features,label)$
        \State initialises $w^{a_{g}}_{0}$, $w^{s}_{0}$ at $t=0$
        \ForAll {communication round $t$}
            $S_{t}\gets$ randomly selected $K \cdot C$ clients
            \ForAll {client $k \in S_{t}$}
        \State $w_{t+1}^{a_{k}} \gets LocalTraining(k,w^{a_{g}}_{t})$
        \Comment{on client $k$}
            \EndFor
        \State $w^{a_{g}}_{t+1} \gets \displaystyle\sum_{k \in S_{t}}
        \frac{n_{k}}{n}w^{a_{k}}_{t+1}$ \Comment{FedAvg on the server}
        \State $X^\prime \gets w^{a_{g}}_{t+1}.encoder(X)$
        \State $D^\prime_{t} \gets (X^\prime,Y)$
        \State $w^{s}_{t+1} \gets CloudTraining(D^\prime_{t},w^{s}_{t})$
        \Comment{on the server}
        \EndFor
        \vspace{0.05cm}
        \hrule
        \vspace{0.05cm}
    \end{algorithmic}
    \caption{Algorithm of semi-supervised FL. $n_{k}$ and $n$ are the numbers of
    unlabelled samples on client $k$ and on all selected clients, respectively.
    $LocalTraining$ is unsupervised learning on the global autoencoder
    $w^{a_{g}}_{t}$ on a client. $CloudTraining$ is supervised learning on the
    classifier $w^{s}_{t}$ on the server.}
    \label{fig:alg}
\end{figure}

In each communication round $t$, the resulting classifier $w^{s}_{t}$ is also
sent to selected clients with the global autoencoder $w^{g_{a}}_{t}$. In order to
locally recognise activities from its observations $X$, a client first uses
the encoder part of $w^{g_{a}}_{t}$ to transform $X$ into its presentation
$X^\prime$, and then feeds $X^\prime$ into the classifier $w^{s}_{t}$ to
recognise the corresponding activities.

%% file: evaluation.tex
\section{Evaluation}
We evaluated our system through simulations on different human activity
datasets with different system designs and configurations. In addition we
evaluated the local activity recognition algorithms of our system on a
Raspberry Pi 4 model B. We want to answer research questions as follow:

\begin{itemize}
    \item Q1. How does our system perform in comparison to supervised
    learning on a centralised server?
    \item Q2. How does our system perform in comparison to semi-supervised FL
    using data augmentation?
    \item Q3. How does our system perform in comparison to supervised FL?
    \item Q4. How do the key parameters of our system, including the
    size of labelled samples on the server and the size of learned
    representations, affect the performance of HAR.
    \item Q5. How efficient is semi-supervised FL on low-cost edge devices.
\end{itemize}

\subsection{Datasets}
We used three HAR datasets that contain time-series sensory data in our
evaluation. The datasets have different numbers of features and activities
with different durations and frequencies.

The Opportunity (Opp) dataset~\cite{Chavarriaga2013} contains short-term and
non-repeated kitchen activities of 4 participants. The Daphnet Freezing of
Gait (DG) dataset~\cite{Bachlin2009} contains Parkinson's Disease patients'
freezing of gaits incidents collected from 10 participants, which are also short-term
and non-repeated. The PAMAP2 dataset~\cite{Reiss2012} contains household and
exercise activities collected from 9 participants, which are long-term and
repeated. The data pre-processing procedure in our evaluation is the same as
described by Hammerla \etal~\cite{Hammerla2016}. Table~\ref{tab:datasets}
shows detailed information about the used datasets after being pre-processed.

\begin{table}[h]
  \caption{HAR datasets in our experiments.}
  \label{tab:datasets}
  \centering
  \begin{tabular}{p{1.5cm}p{3.5cm}p{1.5cm}p{1.5cm}p{1cm}p{1cm}}
    \toprule
    \textbf{Dataset} & \textbf{Activities} & \textbf{Features}
    & \textbf{Classes} & \textbf{Train} & \textbf{Test}\\
    \midrule
    Opp & Kitchen & 79 & 18 & 651k & 119k\\
    DG & Gait & 9 & 3 & 792k & 81k\\
    PAMAP2 & Household \& Exercise & 52 & 12 & 473k & 83k\\
    \bottomrule
  \end{tabular}
\end{table}

\subsection{Simulation setup}
We simulated a semi-supervised FL that runs unsupervised learning on 100
clients to locally update autoencoders and runs supervised learning on a
server to update a classifier. In each communication round $t$, the server
selects $100\cdot C$ clients to participate in the unsupervised learning, and
$C$ is the fraction of clients to be selected. Each selected client uses
its local data to train the global autoencoder $w^{a_{g}}_{t}$ with a learning
rate $lr_{a}$ for $e_{a}$ epochs. The server conducts supervised learning to
train the classifier $w^{s}_{t}$ with a learning rate $lr_{s}$ for $e_{s}$
epochs. For each individual simulation setup, we conducted 64 replicates with
different random seeds.

Based on the assumption that a server is more computationally powerful than a
client in practice, we set the learning rates $lr_{a}$ and $lr_{s}$ as 0.01
and 0.001, respectively. Similarly, we set the numbers of epochs $e_{a}$ and
$e_{s}$ as 2 and 5, because an individual client is only supposed to run a
small number of epochs of unsupervised learning and a server is capable of
doing more epochs of supervised learning. The reason for setting $e_{s}=5$ is
to keep the execution time of our simulation in an acceptable range.
Nevertheless, we believe that this parameter on the server can be set as a
larger number in real-world applications where more powerful clusters and
graphics processing units (GPUs) can be deployed to accelerate the convergence
of performance.

\subsubsection{Baselines}
To answer Q1 and Q2, we consider two baselines to 1) evaluate whether the
autoencoders in our system improve the performance of the system and 2)
compare the performance of our system to that of data augmentation based
semi-supervised FL.

Since we assume that labelled data exist on the server of the system, thus
for ablation studies, we consider a baseline system that only uses these
labelled data to conduct supervised learning on the server and sends trained
models to clients for local activity recognition. This system trains an LSTM
classifier on labelled data on the server and does not train any autoencoders
on clients. We refer to this baseline of a centralised system as \textbf{CS}.
Comparing the performance of CS to that of our proposed system will indicate
whether the autoencoders in our system have any effectiveness in improving
the performance of the trained model.

To compare our system with the state of the art, we consider a
semi-supervised FL system that uses data augmentation to generate pseudo
labels as another baseline. We refer to this baseline as \textbf{DA}. It
first conducts supervise learning on labelled data on the server to train an
LSTM classifier. It then follows standard FL protocols to sends the trained
global model to clients. Each client uses the received model to generate
pseudo labels on their unlabelled local data. To introduce randomness in data
augmentation, we feed sequences with randomised lengths into the model when
generating labels. The sequences are then paired with the labels that are
generated from them as a pseudo-labelled local dataset, which is used for
locally updating the global model.

\subsubsection{Autoencoders and classifiers}
We implement three schemes for our system with different autoencoders,
including simple autoencoders, convolutional autoencoders, and
LSTM-autoencoders.

The first scheme uses a simple autoencoder with fully connected
(FC) layers to learn representations from individual samples in
unsupervised learning and uses a classifier that has an LSTM cell with its
output hidden states connected to an FC layer and a Softmax layer, which we
refer to as \textbf{FC-LSTM}.

The second scheme uses 1-d convolutional and transposed convolutional layers
in its autoencoder. The convolutional layer has 8 output channels with kernel
size 3 and has both stride and padding sizes equal to 1. The output is batch
normalised and then fed into a ReLU layer. To control the size of the encoded
$h$, after the ReLU layer, we flatten the output of 8 channels and feed it
into a fully connected layer that transforms it into the $h$ with a specific
size. For the decoder part, we have a fully connected layer whose output is
unflattened into 8 channels. Then we use a 1-d transposed convolutional layer
that has 1 output channel with kernel size 3 and has both stride and padding
sizes equal to 1, to generate the decoded $X^\prime$. The LSTM classifier of
this scheme has the same structure as that in FC+LSTM. We refer to this
scheme as \textbf{CNN-LSTM}.

For the third scheme, we use an LSTM-autoencoder to capture time-series
information in local unsupervised learning. Both the encoder and the decoder
have 1 LSTM cell. It uses a Softmax classifier that has a fully connected
layer and a Softmax layer. We refer to this scheme as \textbf{LSTM-FC}

For the LSTM classifiers in our experiments, we adopted the bagging (\ie,
bootstrap aggregating) strategy similar to Guan and
Pl{\"{o}}tz~\cite{Guan2017} to train our models with random batch sizes and
sequence lengths. In all schemes, we used the mean square error (MSE) loss
function for autoencoders and the cross-entropy loss function for
classifiers. We used the Adam optimiser in the training of all models. All
the deep learning components in our simulations were implemented using
PyTorch libraries~\cite{pytorch}.

\subsubsection{Label ratio and compression ratio}
\label{sec:label}
We adjusted two parameters to control the amount of labelled data and the
size of representations. For an original training dataset that has $N^{l}$
time-series samples with labels, we adjusted the \emph{label ratio}
$r^{l}\in(0,1)$ and took $r^{l}\cdot N^{l}$ samples from it as the labelled
training dataset on the server. Since the samples are formed as time-series
sequences, to avoid breaking the activities by directly taking random samples
from the sequences, we first divided the entire training set into 100
divisions. We then randomly sampled $100\cdot r^{l}$ divisions and
concatenated them as the labelled training dataset on the server. For a
training dataset whose observations have $N^{f}$ features, we adjusted the
\emph{compression ratio} $r^{f}\in(0,1)$ and used the rounded value of
$r^{f}\cdot N^{f}$ as the size of the representation when training
autoencoders.

\subsubsection{IID and Non-IID local data}
We used two strategies to generate local training data for clients from 
training datasets. In both strategies, the number of allocated
samples for each client, \ie, $n_{k}$, equals to $\frac{n^{o}}{n^{p}}$, where
$n^{o}$ is the number of samples in the original training dataset shown in
Table~\ref{tab:datasets} and $n^{p}$ is the number of participants (e.g., 4
for the Opp dataset) of the original dataset.

To generate \textbf{IID local training datasets}, we divided the 
training dataset into 100 divisions. For a client to be allocated $n_{k}$
samples, its local training data evenly distribute in these divisions. In
each division, a time window that contains $\frac{n_{k}}{100}$ continuous
samples is randomly selected, without their labels, as the client's sample
fragment in this division. The sample fragments from all divisions are then
concatenated as the local training dataset for the client in the IID
scenario.

For \textbf{Non-IID local training datasets}, we randomly located a time
window with length $n_{k}$ in the training dataset and used the samples
without labels in the time window as the local training dataset for the
client in the Non-IID scenario. By this means, the local training dataset of
each client can only represent the distribution within a single part in the
unlabelled dataset.

\subsection{Edge device setup}
Apart from simulations, to answer Q5, we evaluated the local activity
recognition part of our system on a Raspberry Pi 4 Model B. The
specifications of the device are shown in Table.~\ref{tab:pi}.

\begin{table}[h]
  \caption{System specifications of Raspberry Pi 4 Model B.}
  \label{tab:pi}
  \centering
  \begin{tabular}{llp{1cm}p{4cm}}
    \toprule
    \textbf{CPU} & Quad core Cortex-A72 (ARM v8) 64-bit SoC @ 1.5GHz\\
    \midrule
    \textbf{RAM} & 4GB LPDDR4-3200 SDRAM\\
    \midrule
    \textbf{Storage} & SanDisk Ultra 32GB microSDHC Memory Card\\
    \midrule
    \textbf{OS} & Ubuntu Server 19.10\\
    \bottomrule
  \end{tabular}
\end{table}

Compared with supervised FL, on the one hand, our system introduces local
autoencoders that encode samples into representations before feeding them
into classifiers, which costs additional processing time. On the other hand,
encoded representations have smaller sizes than original samples do,
which reduces the processing time of classifiers. To understand how these two
factors affect the overall local processing time, we tested both supervised FL and
our system on the Raspberry Pi and compared their performances. We divided
the testing datasets into one-second-long sequences and measured the overall
processing time of the trained models (\ie, autoencoders + classifiers) on
each sequence, in order to calculate the overhead for each one-second time
window.

\subsection{Metrics}
We evaluated the performance of the global autoencoder and the classifier
with the testing datasets at the end of every other communication round. We
first used a time window to select 5000 samples each time. As the sampling
frequency in the processed datasets is approximately $33Hz$, this time window
represents activities in about 2.53 minutes. We then applied the global
autoencoder on the samples in the time window to encode them into a sequence
of labelled representations. The classifier was applied to the sequence
of representations to recognise the activities, which were then compared with
the ground truth labels. We calculate the \emph{accuracy} in the time window,
which is the fraction of correctly classified representations among all
representations. The accuracies from different time windows are averaged as
the accuracy of the system. In every other communication round $t$, we calculate
the average value from 64 simulation replicates and its standard error.

%% file: results.tex
\section{Results}
We find that our proposed semi-supervised FL system has higher accuracy than
the centralised system that only conducts supervised learning on the server.
The accuracy is also higher than that of data augmentation based
semi-supervised FL and is comparable to that of supervised FL that requires
more labelled data and bigger local models. In addition, it has marginal
local activity recognition time on a low-cost edge device.

\subsection{Analysis of autoencoders and classifiers}
We first look at the contribution of the autoencoders in our proposed system.
As in our assumption, the server has some labelled data that can be used for
supervised learning. If the server has enough labelled data to train a decent
model, its accuracy may be higher than that of a semi-supervised FL. Thus the
centralised system (CS) is a natural baseline that our system needs to
surpass.

We adjust the label ratio $r^l$ from $1/2$ to $1/32$ in ablation studies
for each scheme and try to find out if our system has higher accuracy than
CS. We find that the scheme FC-LSTM (\ie, using simple autoencoders and LSTM
classifiers) has lower accuracy than the CS baseline does under all circumstances.
Therefore we remove it from our analysis. For the other schemes, we keep two
$r^l$ values that lead to the two highest accuracies that are higher than
that of the CS baseline on each dataset. Thus on the Opp dataset, our schemes have
better performance than the CS baseline does when $r^l=\{1/16, 1/32\}$. On
the DG and PAMAP2 datasets, we have $r^l=\{1/2, 1/4\}$. We test our schemes
on both IID and Non-IID data but have not found significant differences in the
accuracy because our schemes do not use any labels locally. Thus we only show
the results on IID data. All schemes' accuracy converges after 50
communication rounds and we only show the results during this period.

\subsubsection{Ablation study of CNN autoencoder}
\begin{figure*}[!t]
    \includegraphics[width=\linewidth]{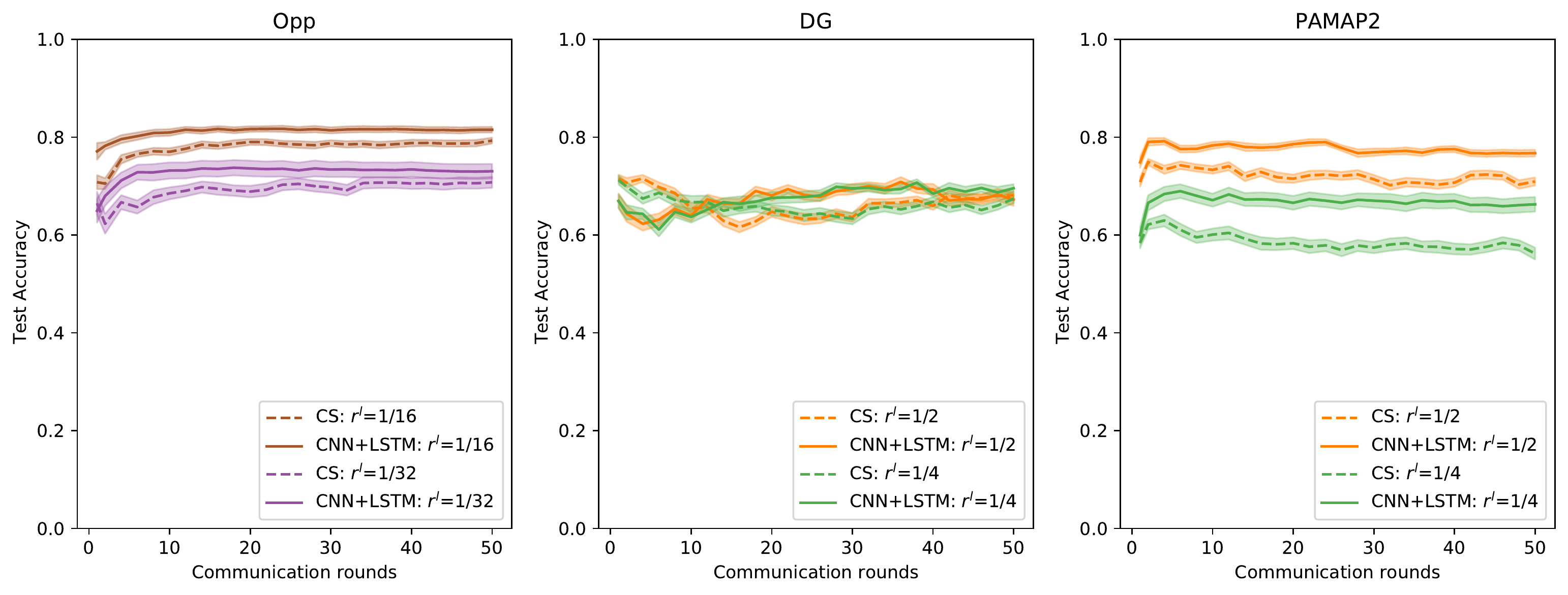}
    \caption{Test accuracy of CNN autoencoders with LSTM classifiers
    (CNN+LSTM), and a centralised system (CS) using LSTM classifiers without
    autoencoders. $r^f=1/2$ for both schemes. CNN+LSTM has higher converged
    accuracy than CS, which means that unsupervised learning on CNN
    autoencoders helps improve the performance.}
    \label{fig:cnn_ablation}
\end{figure*}

Fig.~\ref{fig:cnn_ablation} shows the accuracy of the scheme CNN+LSTM and
the scheme CS, with $r^f=1/2$ on different datasets. As the round of
communications increases, the accuracy of all schemes goes up and converges.
The converged accuracy of CNN+LSTM schemes, \ie, using a convolutional
autoencoder to learn representations locally and using an LSTM classifier for
supervised learning in the cloud, is higher than that of the CS schemes that
only conduct supervised learning in the cloud. This means that training 
CNN autoencoders locally indeed contributes to improving the accuracy of the
system. When $r^l$ decreases, the converged accuracy of CNN+LSTM goes down on
all datasets, which means that it is sensitive to the change of label ratios.

\subsubsection{Ablation study of LSTM autoencoder}
\begin{figure*}[!t]
    \includegraphics[width=\linewidth]{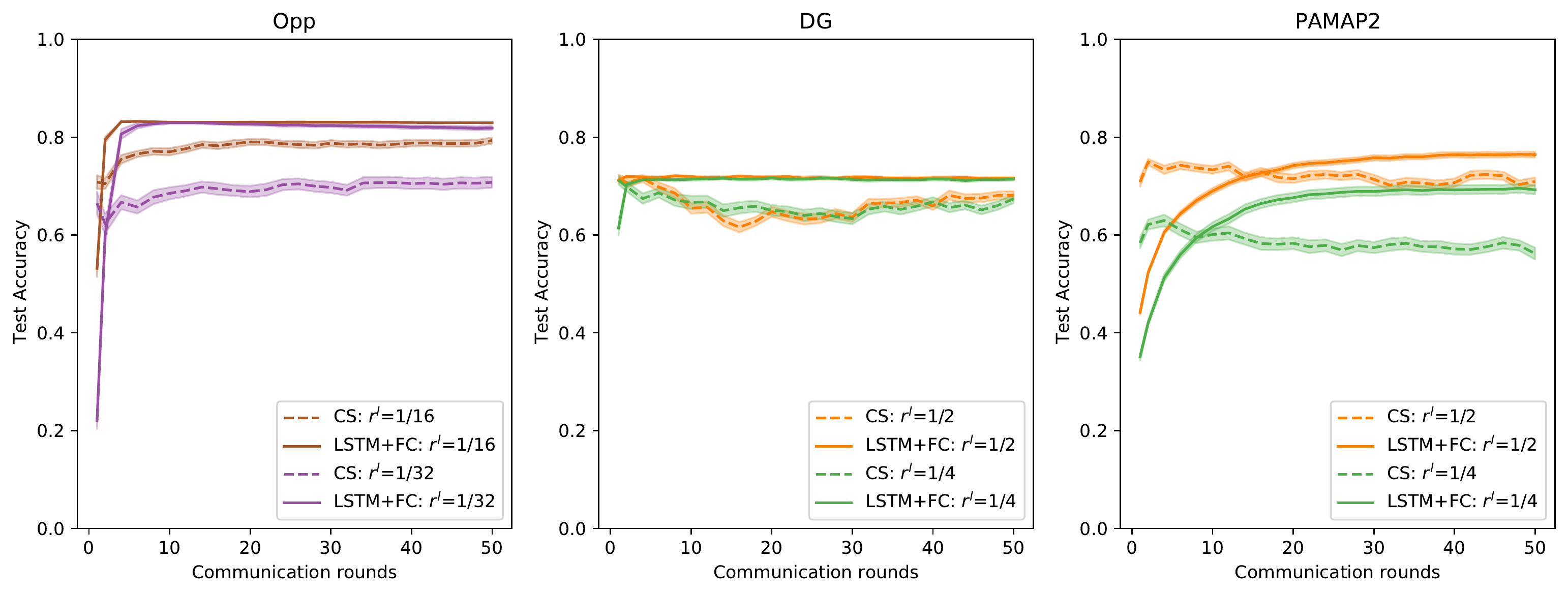}
    \caption{Test accuracy of LSTM autoencoders with Softmax classifiers
    (LSTM+FC), and a centralised system (CS) using LSTM classifiers without
    autoencoders. $r^f=1/2$ for both schemes. LSTM+FC has higher converged
    accuracy than CS. It is less sensitive to the change of $r^l$ than
    CNN+LSTM on the Opp and DG datasets.}
    \label{fig:lstm_ablation}
\end{figure*}

Fig.~\ref{fig:lstm_ablation} shows the accuracy of the scheme LSTM+FC and the
scheme CS, with $r^{f}=1/2$. It demonstrates similar trends as
Fig.~\ref{fig:cnn_ablation} does. The accuracy of LSTM+FC, \ie, using LSTM
autoencoders locally and using Softmax classifiers for supervised learning in
the cloud, is higher than that of CS that runs centralised and supervised
learning without using unlabelled local data. However, LSTM+FC is less
sensitive to the change of label ratios. For example, its converged accuracy
on the Opp dataset is almost the same when we change $r^l$ from $1/16$ to
$1/32$. This would enable us to achieve similar performance but require fewer
labelled data compared to CNN+LSTM.

The experimental results show that, when implementing a semi-supervised FL
system for HAR, both CNN autoencoders and LSTM autoencoders can improve the
accuracy of the system. Using LSTM autoencoders is less sensitive to the
change of available labelled data in the cloud. In the rest of our analyses
of our results, we only show the accuracy of the LSTM+FC scheme.

\subsection{Comparison with different FL schemes}
We now analyse the performance of our system in comparison with
semi-supervised FL using data augmentation (DA) to generate pseudo labels and
supervised FL having labelled data available on clients.

\subsubsection{Comparison with DA}
\begin{figure*}[!t]
    \includegraphics[width=\linewidth]{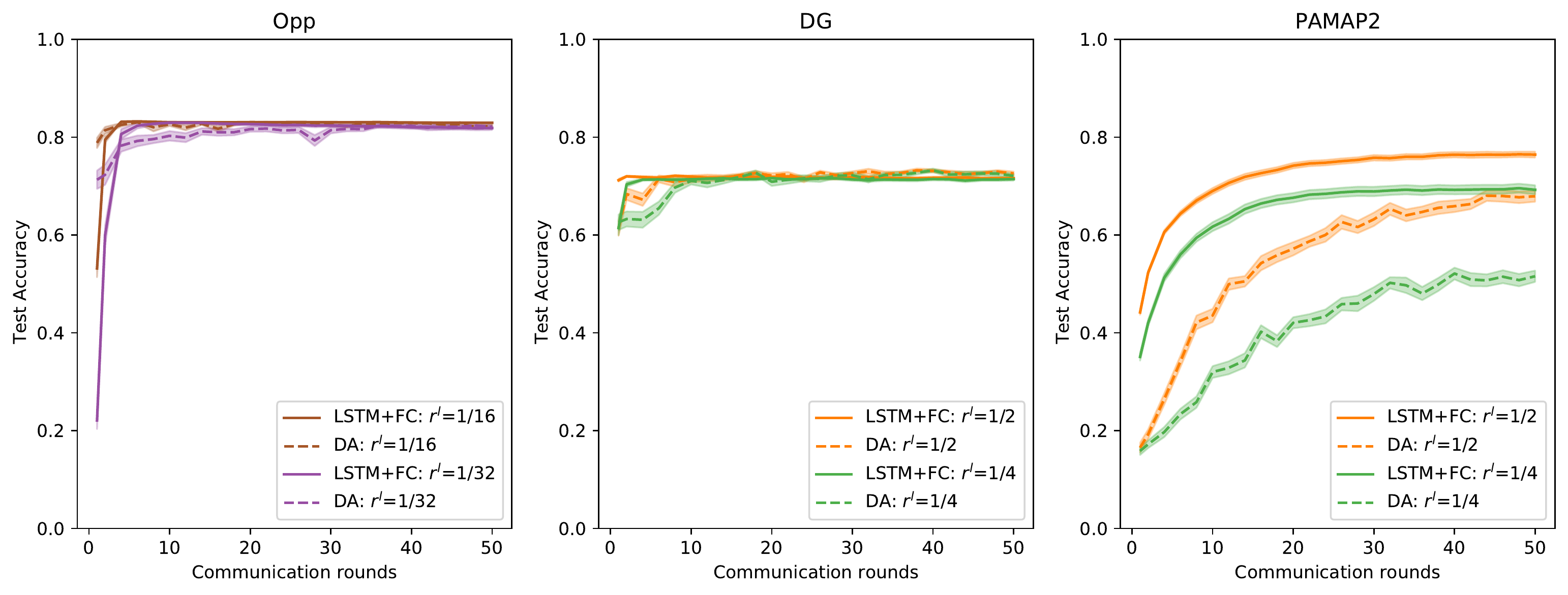}
    \caption{Test accuracy of LSTM autoencoders with Softmax classifiers
    (LSTM+FC), and semi-supervised FL using data augmentation (DA). There is
    not significant difference between their converged accuracy on the Opp
    and DG datasets. LSTM+FC has higher converged accuracy than DA on the
    PAMAP2 dataset.}
    \label{fig:fake_label}
\end{figure*}

Fig.~\ref{fig:fake_label} shows the accuracy of both LSTM+FC and DA on three
datasets. On the Opp and DG datasets, the accuracy of DA increases more
slowly than LSTM+FC does. But once the accuracy of both schemes converge,
they do not show significant differences. On the PAMAP2 dataset, the
converged accuracy of LSTM+FC is higher than that of DA. We also find that,
although the accuracy of DA on the Opp and DG datasets is higher than that of
CS in Fig.~\ref{fig:lstm_ablation}, its accuracy on the PAMAP2 dataset in
Fig.~\ref{fig:fake_label} converges more slowly than CS does in
Fig.~\ref{fig:lstm_ablation}. This indicates that using the received global
LSTM model to generate pseudo labels and then training the model on these
pseudo labels may damage the testing accuracy. Although we used time-series
sequences with randomised lengths to generate pseudo labels in our
experiments, DA may still risk overfitting the model to the training data and
consequently has slower speed to achieve decent accuracy on testing data.

Our results indicate that, for semi-supervised FL, using locally trained
autoencoders can achieve higher converged accuracy than using data
augmentation to generate pseudo labels. In addition, compared with DA, our
scheme is independent from the specific tasks provided by the server. For
example, if one client uses its unlabelled data to access multiple FL
servers that conduct different tasks, with data augmentation, the client has
to generate pseudo labels for each of models of these tasks. In our scheme,
the client only conducts unsupervised learning locally using unlabelled data
to learn general representations, which is independent from the labels in the
cloud.

\subsubsection{Comparison with supervised FL}
\begin{figure*}[!t]
    \includegraphics[width=\linewidth]{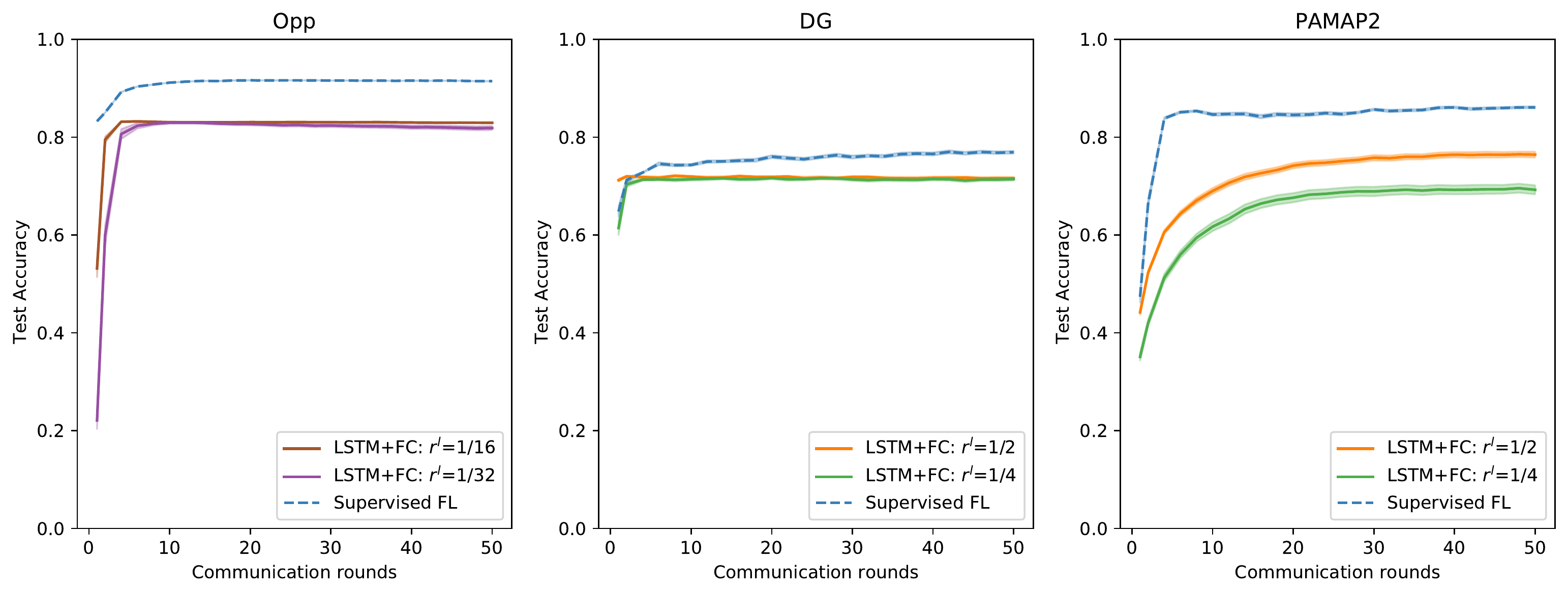}
    \caption{Test accuracy of LSTM autoencoders with Softmax classifiers
    (LSTM+FC, $r^f=1/2$), and supervised FL using 100\% features and labels. The
    converged accuracy of LSTM+FC is comparable to that of the supervised FL
    and requires fewer labelled data.}
    \label{fig:supervised}
\end{figure*}

Fig.~\ref{fig:supervised} shows the accuracy of LSTM+FC with $r^f=1/2$ and a
supervised FL scheme. The supervised FL uses all the information (\ie, 100\%
features and 100\% labels) in the training datasets. Therefore it has higher
accuracy than that of LSTM+FC. However, our scheme enables a trade-off
between the performance of the system (\ie, accuracy), the cost of data
annotation (\ie, label ratio), and the size of models (\ie, compression
ratio). For example, having larger compression ratio $r^f=3/4$ on the PAMAP2
dataset can lead to a higher accuracy (shown in
Fig.~\ref{fig:compression_ratio}) that is comparable to that of the
supervised FL.

The experimental results suggest that we can implement FL systems in a
semi-supervised fashion with fewer needed labels than those in supervised FL,
meanwhile achieve comparable accuracy. Although one of the motivations of FL
is to hold models instead of personal data in the cloud to address potential
privacy issues, the data held by the server of our system do not have to be
from the users of the service of the system. This kind of dataset in the
cloud has been used in FL to address other challenges such as dealing with
Non-IID data by creating a small globally shared dataset ~\cite{Kairouz2019}
and does not necessarily contain private information. We believe that service
providers can collect these data from open datasets, or from laboratory
trials in controlled environment where data subjects give their consents to
contribute their data.

\subsection{Analysis of compression ratio}
\label{sec:compression}
We also investigate how the compression ratio $r^{f}$ of autoencoders affects
the accuracy of our system. It is an important factor that can affect the
number of parameters and the size of local models. These local models are
regularly uploaded from the clients to the server over network, hence their
sizes affect the outbound traffic. We use $r^{f}=\{3/4,1/2, 1/4,1/8\}$ on all
datasets. We keep $r^{l}=1/16$ for the Opp dataset and $r^{l}=1/2$ for both
the DG and PAMAP2 datasets.

\begin{figure*}[t!]
    \includegraphics[width=\linewidth]{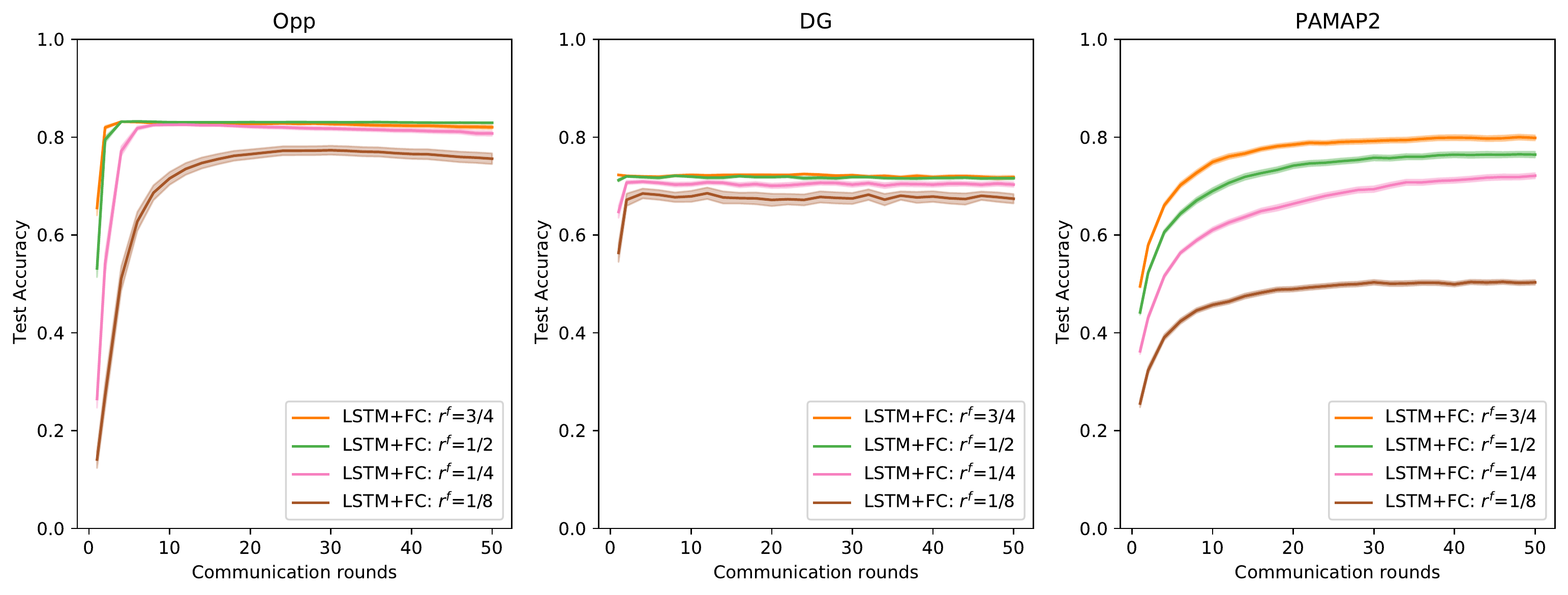}
    \caption{Test accuracy with different compression ratios
    $r^{f}=\{3/4,1/2,1/4,1/8\}$. $r^l=1/16$ for OPP and $r^l=1/2$ for both DG
    and PAMAP2. The accuracy on OPP and DG is not affected much when changing
    $r^f$ from $3/4$ to $1/4$. PAMAP2 is more sensitive to the change of
    $r^f$ than the other two datasets.}
    \label{fig:compression_ratio}
\end{figure*}

Fig.~\ref{fig:compression_ratio} demonstrates that, on the Opp and the DG
datasets, our system can compress an original sample into a representation
whose size is only $1/4$ of the original sample without significantly
affecting the accuracy. On the PAMAP2 dataset, increasing $r^f$ from $1/2$ to
$3/4$ can lead to accuracy that is comparable to that of the supervised FL
scheme.

Changing the compression ratio in our system allows us to exchange accuracy
with model sizes, or \textit{vice versa.} When used data are not sensitive to
the compression ratio (\eg, Opp and DG), compressing samples into smaller
representations may significantly reduce the size of local models that are
uploaded from the clients to the server, which may lead to lower network
traffic.

\subsection{Running time at the edge}
\begin{figure}[!t]
   \includegraphics[width=.75\linewidth]{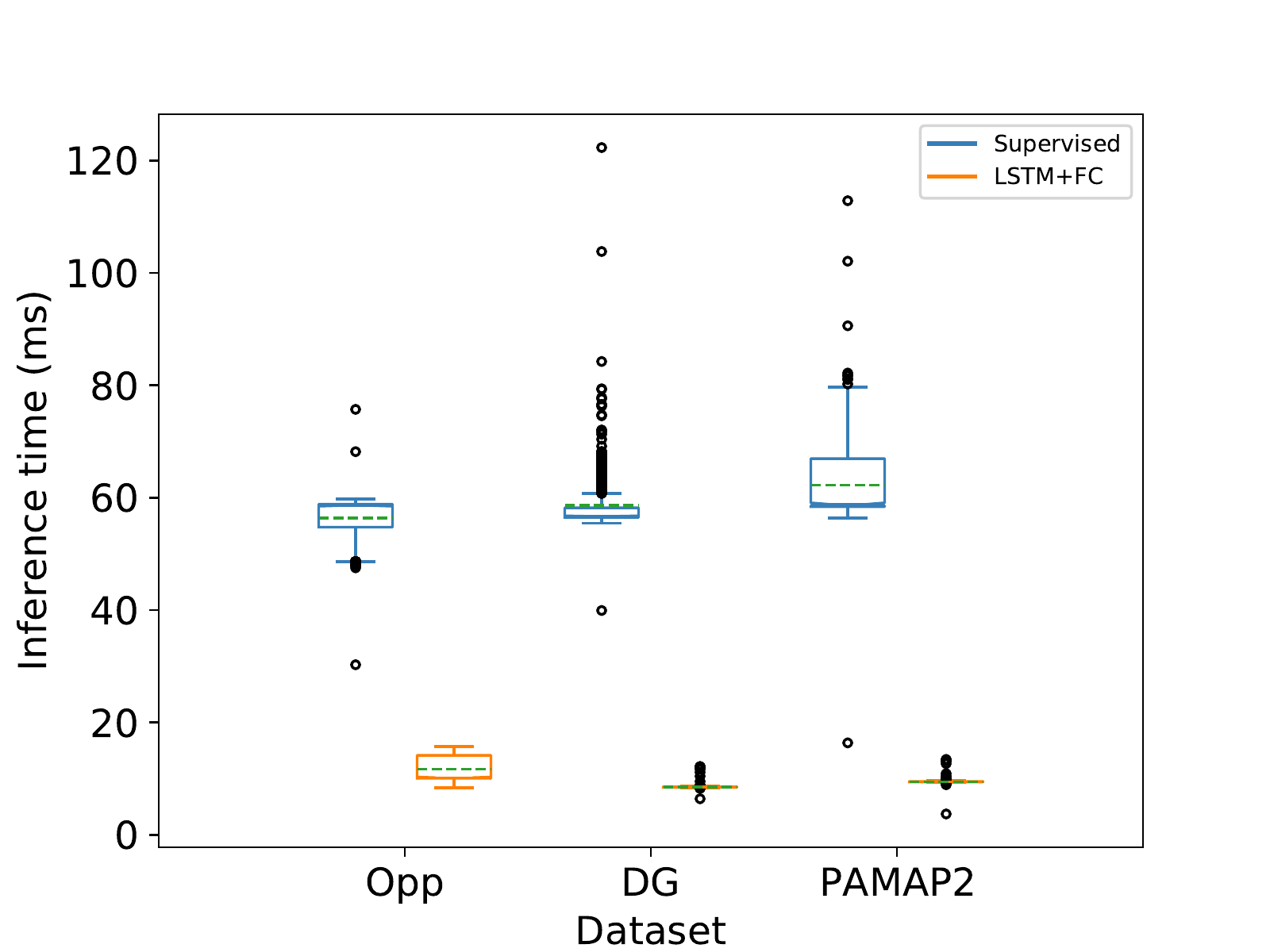} 
   \caption{Local running time on a Raspberry Pi 4 Model B. Dashed lines are
   average values. Each data point is the processing time when recognising
   activities of a one-second long sequence. The running time of our
   system on all three datasets is significantly ($p<0.001$) lower than that
   of the supervised FL.}
   \label{fig:pi}
\end{figure}

We evaluated the local activity recognition using both supervised FL 
and our system (LSTM+FC) with $r^{f}=0.5$ on a Raspberry Pi. As shown in
Fig.~\ref{fig:pi}, the processing time of our system is significantly lower
($p<0.001$) than that of supervised FL on all datasets. Although the
autoencoder in our system inevitably causes extra processing time as it
increases the length of the local pipeline, the LSTM cell in the autoencoder
encodes the input data into smaller representations. In contrast, the LSTM
classifier in the supervised FL transforms input data into hidden states that
have larger sizes. This reduction of the amount of data leads to a shorter
overall processing time than that of the supervised FL.

Combined with the results of Sec.~\ref{sec:compression}, our experimental
results show that running unsupervised learning on autoencoders can reduce
both the size of local models and the size of data processed by classifiers.
This can potentially improve not only the outbound network traffic, but also
the efficiency of local activity recognition.

%% file: discussion.tex
\section{Discussion}
Our experimental results show that HAR with semi-supervised FL can achieve
comparable accuracy to that of supervised FL. We now discuss how these
results can contribute to the system design of FL systems and possible
research topics.

\subsection{FL servers can do more than FedAvg}
In canonical FL systems, servers only hold global models and use the FedAvg
algorithm to aggregate received local models into new global models. This
design consideration is due to the privacy concerns of having personal data
on the servers. Our findings suggest that running supervised learning with a
small amount of labelled data on the servers can alleviate individual users
from labelling their local data. Therefore, we suggest that FL systems may
consider maintaining datasets that do not contain private information on their
servers to support semi-supervised learning. Apart from implementing the
FedAvg algorithm in every communication round, servers can conduct more
epochs of supervised learning than individual clients can do, since they have
more computational resources and fewer power constrains than clients
do. This can help the performance of the models converge faster.

\subsection{Learning useful representations, not bias}
By training autoencoders locally, semi-supervised FL is not affected by
Non-IID data because it does not use any labels locally. This sheds light on
a new solution, which is different from data augmentation or limiting
individual contributions from clients~\cite{Kairouz2019}, to address the
Non-IID data issue in FL. Although, our work focuses on semi-supervised FL
where no labels are available on clients, we suggest that supervised FL can
also consider learning general representations apart from the mappings from
features to labels, and use the learned representations to help alleviate the
bias caused by Non-IID data.

Another possible application of semi-supervised FL is to defend against
malicious users who attack the global model through data
poisoning~\cite{Kairouz2019}. Labels of local data are a common attack vector
in FL. Adversaries can manipulate (\eg, flipping) the labels in their local
data to affect the performance of their local models, thereby affecting the
performance of the global model. Such an attack will be removed if we do not
use local labels. We suggest that security researchers in FL should consider
semi-supervised FL as a possible scheme to defend from data poisoning
attacks.

\subsection{Smaller models via unsupervised learning}
In supervised FL, as the complexity of an ML task goes up, the size of the
model of the task increases. This eventually leads to increasing numbers of
parameters and increasing network traffic when uploading local models to the
server. Semi-supervised FL only uploads trained autoencoders from clients to
the server. Our experimental results suggest that the performance of
semi-supervised FL can still converge to an acceptable level even if we use
high compression rates, which help with reducing a significant amount of
model parameters. The compression rate can be considered as a system
parameter that can be tuned to reduce the size of models, as long as the key
representations can be learned and the performance of the system can be
guaranteed. This makes our system suitable in scenarios that have demanding
network conditions. Although in this paper, we only focus on the application
of HAR, such effects may exist in other applications with different types of
data, which should be further investigated.

%% file: conclusions.tex
\section{Conclusions}
HAR using IoT sensory data and FL systems can empower many real-world
applications including processing the daily activities and the changes to
these activities in people living with long-term conditions. The difficulty
of obtaining labelled data from end users limits the scalability of the FL
applications for HAR in real-world and uncontrolled environments. In this
paper, we propose a semi-supervised FL system to enable HAR in IoT
environments. By training LSTM autoencoders through unsupervised learning on
FL clients, and training Softmax classifiers through supervised learning on
an FL server, our system can achieve higher accuracy than centralised systems
and data augmentation based semi-supervised FL do. The accuracy is also
comparable to that of a supervised FL system but does not require any locally
labelled data. In addition, it is has simpler local models with smaller size
and faster processing speed.

Our future research plans are to investigate fully unsupervised FL systems
that can support anomaly detection through analysing the difference between
local and global models. We believe that such systems will enable many useful
real-time applications in HAR where useful labels are rare or extremely
difficult to collect.